# Exploring Recurrent-Depth Language Models for Multitask Clinical Text Processing

Xi Yang, PhD
*Advanced Intelligence and Research*
*Eli Lilly and Company*
Indianapolis, IN, USA
xi.yang5@lilly.com

Guodong Liu, PhD
*Advanced Intelligence and Research*
*Eli Lilly and Company*
Indianapolis, IN, USA
guodong.liu@lilly.com

Cheng Peng, PhD
*Dept. of Health Outcomes and Biomedical Informatics*
*University of Florida*
Gainesville, FL, USA
c.peng@ufl.edu

Kaleb E. Smith, PhD
*Higher Education and Research*
*NVIDIA*
Santa Clara, CA, USA
kasmith@nvidia.com

Yingying Zhang
*Dept. of Computer Science*
*University of Texas Rio Grande Valley*
Edinburg, TX, USA
yingying.zhang01@utrgv.edu

Xing He, PhD
*Dept. of Biostatistics and Health Data Science*
*Indiana University*
Indianapolis, IN, USA
hexing@iu.edu

Yi Guo, PhD
*Dept. of Health Outcomes and Biomedical Informatics*
*University of Florida*
Gainesville, FL, USA
yiguo@ufl.edu

Jiang Bian, PhD
*Dept. of Biostatistics and Health Data Science*
*Indiana University*
Indianapolis, IN, USA
bianji@iu.edu

Haoteng Tang, PhD
*Dept. of Computer Science*
*University of Texas Rio Grande Valley*
Edinburg, TX, USA
haoteng.tang@utrgv.edu

Yonghui Wu, PhD
*Dept. of Health Outcomes and Biomedical Informatics*
*University of Florida*
Gainesville, FL, USA
yonghui.wu@ufl.edu

**Abstract—Clinical natural language processing includes diverse tasks such as information extraction, classification, and question answering. Developing one model for all tasks could simplify deployment and improve knowledge transfer. We evaluated whether small recurrent-depth language models can meet this need through joint fine-tuning and compared their performance with standard transformers. We fine-tuned Ouro-1.4B-Thinking, Ouro-2.6B-Thinking, HRM-Text-1B, Qwen3-1.7B, and Qwen3-8B on 9 datasets covering information extraction, classification, inference, and question answering, and evaluated them on four additional datasets excluded from training. Across these 13 datasets, Ouro-1.4B-Thinking achieved higher performances than Qwen3-1.7B on 11 datasets; Ouro-2.6B-Thinking was competitive with Qwen3-8B on most tasks. Ouro models showed the largest gains on inference and question answering, whereas HRM-Text-1B showed mixed results. These findings support further study of recurrent-depth models for clinical tasks through controlled experiments to isolate architectural effects and investigate how recurrent computation influences reasoning, knowledge acquisition, and generalization.**

***Keywords—clinical natural language processing, latent reasoning, looped transformers, recurrent depth, multitask learning, clinical information extraction, small language models***

## I. Introduction

Clinical narratives in electronic health records (EHRs) document the detailed patient information needed for clinical and translational research such as adverse drug events, patient health status and progression, and social determinants of health. Natural language processing (NLP) is the key technology to extract the information into structured representations [1]. In the past decade, research communities organized the National NLP Clinical Challenges (n2c2) series to establish widely used benchmark datasets for these tasks [2]–[5] and motivated researcher to explore new methods to approach them. Recently, fine-tuned decoder-only large language models (LLMs) have matched or modestly exceeded encoder-based models on clinical concept and relation extraction [6], [7]. Many clinical tasks require more than locating text spans; determining whether a medication was started or stopped, or whether a patient meets a trial eligibility criterion, requires interpreting context. Clinical applications also often require several such tasks together, and a model that supports these tasks with one set of weights could reduce the need to maintain separate task-specific models. Small language models are receiving increasing attention in healthcare [8], but their performance across diverse clinical tasks remains an important consideration when selecting a model.

One approach to improving model performance on complex tasks is to allocate additional computation at inference time. Chain-of-thought (CoT) prompting generates intermediate reasoning steps in natural language before producing the final answer [9], and recent models such as Qwen3 offer an optional thinking mode [10]. Yet, explicit CoT has not consistently improved performance on clinical NLP tasks. A systematic evaluation of 95 LLMs across 87 clinical text understanding tasks found that 86.3% of the models performed worse with CoT than with direct prompting [11]. Similarly, a biomedical benchmark reported that standard prompting outperformed CoT for text classification and named entity recognition [12]. Most clinical NLP datasets provide only task-specific labels without clinician-verified reasoning traces. This motivates evaluation of models that perform additional computation without emitting intermediate reasoning text.

Recurrent-depth language models repeatedly apply shared layers to iteratively refine hidden representations before generating an output, an approach often described as latent reasoning [13], [14]. This form of recurrence differs from segment-level recurrence, which propagates memory across segments of a long input sequence [15], [16]. Saunshi et al.

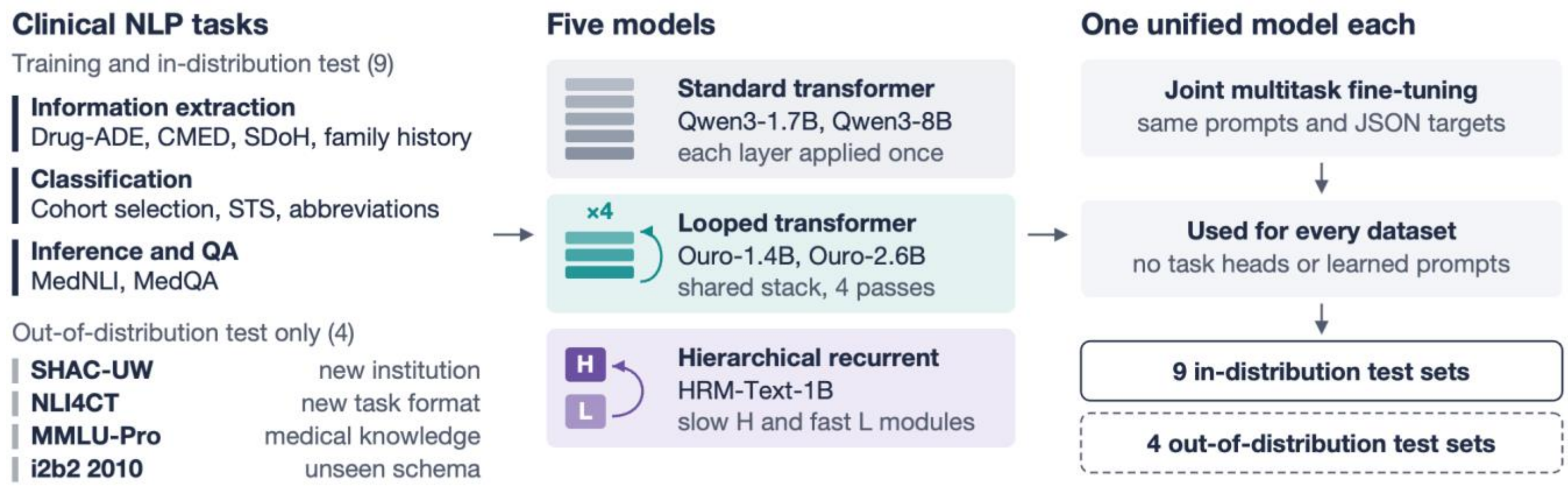

Fig. 1. Study overview. Each model was fine-tuned jointly as one unified multitask model on nine in-distribution clinical NLP datasets spanning four task families and evaluated on their test sets and on four out-of-distribution test sets.

studied latent thoughts in looped transformers [14], and Geiping et al. examined how recurrent steps affect reasoning performance [17]. Building on this paradigm, the Ouro family has performed competitively with larger standard transformer models on general-domain benchmarks [18]. HRM-Text introduces coupled recurrent modules with different update frequencies and has also shown competitive general-domain performance with relatively limited training data [19]. Unlike explicit chain-of-thought prompting or supervision, latent reasoning emerges through recurrent refinement of hidden representations without requiring intermediate reasoning traces. This characteristic is particularly appealing for clinical NLP, where high-quality chain-of-thought annotations are difficult to obtain because clinical reasoning is costly to annotate and often unavailable due to privacy and workflow constraints. However, existing evaluations of recurrent-depth models have focused on mathematics, code, and general-domain question answering [20], and it remains unclear whether their benefits extend to clinical NLP. To our knowledge, this is the first study to systematically evaluate pretrained depth-recurrent language models, including looped and hierarchical recurrent models, across diverse clinical NLP benchmarks under joint multitask fine-tuning.

This study aims to evaluate whether small recurrent-depth language models can support diverse clinical NLP tasks through joint fine-tuning. We compared two recurrent-depth model families, the Ouro looped models at 1.4 and 2.6 billion parameters [18] and the hierarchical recurrent HRM-Text-1B [19], with Qwen3-1.7B and Qwen3-8B [10]. Following our previous study [1], we formulated all tasks as text-to-text generation and fine-tuned one unified multitask model per checkpoint on nine clinical NLP datasets spanning four task families. We evaluated each model on these datasets and four additional datasets excluded from training (Fig. 1). All models generated direct answers without explicit reasoning text. We examined whether recurrent-depth models outperform similarly sized transformers, remain competitive with larger transformers, and transfer across institutions, task formats, knowledge domains, and annotation schemas.

## II. Methods

### A. Datasets and Tasks

We used nine clinical NLP datasets for training and in-distribution evaluation (Table I), spanning four task families. Clinical information extraction aims to identify clinical concepts together with their attributes and relations. We used four benchmarks: medication and adverse drug event (Drug-ADE) extraction from the 2018 n2c2 Track 2 [3], family history extraction from the 2019 n2c2 Track 2 [21], contextualized medication event extraction (CMED) from the 2022 n2c2 Track 1 [4], and social determinants of health extraction from the MIMIC subset of the Social History Annotation Corpus (SHAC-MIMIC) in the 2022 n2c2 Track 2 [5]. Clinical text classification assigns a label to a document or text span. We used cohort selection from the 2018 n2c2 Track 1 [22], which determines whether a patient meets a clinical trial eligibility criterion, clinical semantic textual similarity (STS) from the 2019 n2c2 Track 1 [23], and the UMN clinical abbreviation sense inventory [24] for abbreviation disambiguation. Natural language inference (NLI) determines whether a hypothesis can be inferred from a premise, for which we used MedNLI [25], and medical question answering (QA) selects the correct answer to an examination question, for which we used the US subset of MedQA-USMLE [26].

For STS, we converted the 0 to 5 similarity scores into five ordinal classes by rounding each score to the nearest half point and grouping the resulting values as {0, 0.5}, {1, 1.5, 2}, {2.5, 3}, {3.5, 4}, and {4.5, 5}. All test scores already lie on the half-point grid, whereas 34% of training scores were rounded, so results for this task are not comparable with the original regression leaderboard.

TABLE I. DATASETS, TASK TYPES, AND SPLIT SIZES

| Dataset | Type | Train | Dev | Test | Med. tok |
|---|---|---|---|---|---|
| **In-distribution (training and test)** | | | | | |
| Cohort selection (2018 T1) | Classif. | 2,363 | 263 | 1,118 | 5,159 |
| Drug-ADE (2018 T2) | IE | 273 | 30 | 202 | 5,400 |
| STS (2019 T1) | Classif. | 1,489 | 165 | 412 | 241 |
| Family history (2019 T2) | IE | 88 | 10 | 107 | 907 |
| CMED (2022 T1) | IE | 350 | 50 | 100 | 1,771 |
| SHAC-MIMIC (2022 T2) | IE | 1,316 | 188 | 373 | 1,018 |
| MedNLI | NLI | 11,232 | 1,395 | 1,422 | 113 |
| MedQA-USMLE (US) | QA | 10,178 | 1,272 | 1,273 | 278 |
| UMN abbreviations | Classif. | 19,257 | 4,127 | 4,127 | 202 |
| **Total** | | **46,546** | **7,500** | **9,134** | |
| **Out-of-distribution (test only)** | | | | | |
| SHAC-UW (2022 T2) | IE | — | — | 518 | 910* |
| NLI4CT | NLI | — | — | 500 | 461* |
| MMLU-Pro (Health) | QA | — | — | 823 | 206* |
| i2b2 2010 | IE | — | — | 256 | 5,116* |

Counts are rendered examples (notes for i2b2). Med. tok: median cl100k token length of the rendered example in the train split (* test split). IE: information extraction; Classif.: classification. The Drug-ADE training count is note-level; its 2,954 paragraph-window examples are not included in the table total.

We further evaluated transfer on four datasets excluded from training (i.e., out-of-distribution). SHAC-UW, the University of Washington subset of the 2022 n2c2 Track 2, tests transfer across institutions with the same annotation schema. NLI4CT tests transfer to clinical trial reports and a different inference formulation [27]. The health subset of MMLU-Pro tests transfer to another medical knowledge benchmark with up to ten answer options [28]. Neither MedQA nor MMLU-Pro is included in the reported training data of the three model families, which we confirmed for HRM-Text by examining its released training data, and Qwen3 and Ouro report MMLU-Pro as a held-out evaluation benchmark [10], [18]. The 2010 i2b2/VA corpus tests transfer to an unfamiliar concept, assertion, and relation schema [2]. Because models produced relation arguments as concept text rather than concept identifiers, we report only concept F1, representing a partial evaluation of the 2010 i2b2 task.

We used official training, development, and test partitions where available. Where no development set was provided, we allocated 10% of the source training partition to development and used the remaining 90% for training. Table I lists the resulting example counts. Across the nine datasets, the note-level training total was 46,546 examples; Drug-ADE paragraph windows are excluded from that total and described below. The dataset sizes were highly imbalanced, ranging from 19,257 abbreviation examples to 88 family history notes.

#### *1) Note- and paragraph-level preparation of Drug-ADE*

Drug-ADE presents a challenge because each note contains many annotations in a long clinical narrative. The 505 notes contain 83,869 concepts and 59,810 relations, averaging about 166 concepts and 118 relations per note. The source training partition contains 50,951 concepts and 36,384 relations across 303 notes requiring models to learn many extraction decisions from a limited number of annotated narratives and long input-output sequences and dense annotations make full-note generation difficult. We therefore prepared Drug-ADE at both note and paragraph levels. A rule-based segmentation method produced 2,954 paragraph-window training examples and resulted in the loss of approximately 0.24% of gold relations that crossed segmentation boundaries. For the final Drug-ADE results in Table III, models generated predictions for paragraph windows; we recombined predictions for each note and scored them against the original note-level gold annotations. We did not use paragraph windows for the other datasets because SHAC documents contain shorter sections, CMED attributes may require evidence beyond one paragraph, and cohort selection requires patient-level interpretation of the full record.

### *B. Models*

We evaluated models from three architecture families including standard decoder-only transformer, looped transformer, and hierarchical recurrent transformer (Table II); the latter two are recurrent-depth models.

TABLE II. MODEL CONFIGURATIONS

| Model | Params | Layers × steps | Pretraining tokens |
|---|---|---|---|
| Qwen3-1.7B | 1.7B | 28 × 1 | 36T |
| Qwen3-8B | 8.2B | 36 × 1 | 36T |
| Ouro-1.4B | 1.4B | 24 × 4 | 7.7T |
| Ouro-2.6B | 2.6B | 48 × 4 | 7.7T |
| HRM-Text-1B | 1B | 16 × 8 (H/L) | 40B |

Layers × steps denotes transformer-layer applications per forward pass, not independent learning layers. Ouro recurrent steps are fixed at four in this study; HRM-Text applies 16-layer modules in eight updates per forward pass. Ouro rows are the reasoning-tuned checkpoints (released as Ouro-Thinking), which add supervised fine-tuning on reasoning data. Pretraining tokens are as reported by each model's authors.

#### *1) Standard transformer*

Qwen3-1.7B and Qwen3-8B are dense decoder-only transformers pretrained on approximately 36 trillion tokens [10]. Qwen3-1.7B has 28 layers with a hidden size of 2,048, and Qwen3-8B has 36 layers and 8.2 billion parameters. Each layer is applied once per token. Qwen3 supports an optional thinking mode that generates explicit reasoning before the answer; we disabled it in all experiments.

#### *2) Looped transformer: Ouro*

Ouro is a family of looped language models in which a single decoder stack is applied repeatedly to the same hidden states [18]. Ouro-1.4B uses a 24-layer stack with a hidden size of 2,048, and Ouro-2.6B is obtained by upcycling the 1.4B model to a 48-layer stack. Both were pretrained on 7.7 trillion tokens with a learned exit gate for adaptive depth; the released models use at most four recurrent steps. We fixed the number of recurrent steps at four during fine-tuning and inference, the depth at which the performance of Ouro peaks [18]. We did not use the exit gate in our experiments. Instead, the same hidden states were refined through four recurrent iterations for every token during fine tuning and inference. Ouro models released two versions at each scale including a base version obtained from a four-stage training process with large scale raw text data as well as 20+ open-sourced SFT datasets, and a reasoning-tuned version, which is an extension of the base model by further training on approximately 8.3 million reasoning-oriented examples in math, code, and science domains [18]. In this study, we mainly explored the thinking version. For naming convention, we will use Ouro-1.4B-T and Ouro-2.6B-T for thinking version for the rest of the paper.

#### *3) Hierarchical recurrent transformer: HRM-Text*

HRM-Text-1B is a hierarchical recurrent language model that performs latent reasoning through two coupled recurrent modules operating at different timescales: a slow high-level module for planning and a fast low-level module for execution [19]. Each module is a 16-layer transformer with a hidden dimension of 1,536. During each forward pass, the model performs two high-level cycles, each consisting of three low-level updates followed by one high-level update, while reusing the parameters of each module across recurrent updates. It was trained from scratch on 40 billion unique tokens of instruction-response pairs, using a task-completion objective with prefix language-model masking, where instruction tokens were processed with bidirectional attention. Training uses truncated backpropagation through recurrent steps (bp); following the original implementation, we fine-tuned the model with bp = 5, the final horizon used during pretraining [19].

### C. Text-to-Text Formulation and Joint Fine-Tuning

We formulated all tasks as text-to-text generation with JSON outputs. To distinguish different tasks, we constructed the system prompt with a two-level task tag in the form of "<<Category: Task>>" followed by an instruction shared by all tasks in that category. In the user prompt, we included the dataset-specific instruction, which defines the label set and JSON output schema, followed by the clinical context. For extraction tasks, the instructions also specified the definition of entities, attributes, and relations. For outputs, we used the JSON format for all tasks to facilitate the downstream post-processing and evaluation.

For the multitask setting, we merged the training sets from all included datasets and jointly fine-tuned each model as a single text-to-text model. The objective was to train one unified model capable of performing diverse clinical NLP tasks using a shared set of parameters. Following our previous work [1], tasks were distinguished only by the task tag and natural-language instruction in the prompt, without task-specific output heads, adapters, learned prompts, or other task-specific parameters. This unified setting reflects a practical deployment scenario in which a single model supports multiple clinical NLP tasks while sharing its capacity across them.

### D. Two-Stage Curriculum Learning

End-to-end information extraction requires a model to identify entities, determine relationships and attributes, and generate structured output. To test whether learning simpler extraction tasks first improves end-to-end performance, we evaluated a two-stage curriculum strategy. In Stage 1, models were trained for three epochs on entity, relation, and attribute subtasks derived from Drug-ADE and CMED annotations; gold entities were provided as input for relation and attribute subtasks. In Stage 2, models from stage 1 were further trained on the merged multitask set using the standard end-to-end formulation. We compared curriculum trained models with models trained directly on the standard multitask objective. Because curriculum models received additional training before multitask fine-tuning, this comparison does not isolate the effect of subtask ordering from increased training exposure.

### E. Experiments and Evaluation

All models were fine-tuned with full-parameter updates using the standard supervised fine-tuning objective, next-token cross-entropy on the target, implemented with a customized Hugging Face trainer [29] in PyTorch [30]. Model-specific wrappers exposed the four-step recurrent forward pass of Ouro and the hierarchical forward pass of HRM-Text as single training steps, so that loss computation, batching, and checkpointing were shared across models. We used fully sharded data parallelism [31] on four NVIDIA B300 GPUs and trained each model for 10 epochs with a maximum sequence length of 32,768 tokens, using bfloat16 (bf16) precision for both training and inference. Optimization hyperparameters followed the supervised fine-tuning settings reported in the original paper of each model [10], [18], [19]. The checkpoint of each model was selected based on its performance on the development sets of the nine training datasets; the four out-of-distribution test sets were never used in training. Qwen3 and Ouro were served with vLLM [32], with decoding constrained to the JSON schema of each task. HRM-Text, whose prefix language-model attention requires custom attention masks in standard serving frameworks [19], was run with PyTorch generation, and its outputs were validated as JSON.

We report micro-F1 for cohort selection; entity-only and end-to-end relation F1 for Drug-ADE; F1 for family history classification; and trigger, attribute or argument, and combined F1 for CMED and SHAC. We report accuracy for STS, MedNLI, MedQA, UMN abbreviation disambiguation, NLI4CT, and MMLU-Pro, and concept F1 for 2010 i2b2. STS uses the five-class formulation described above rather than the original regression score. NLI4CT uses the training-time three-label prompt on its two-label test set. Official scripts were used without modification for cohort selection and family history; the remaining extraction tasks required conversion of generated text to offsets as described below. Per-dataset performance was the primary outcome. Because each configuration was trained once, differences are descriptive and do not establish statistical significance or equivalence.

For the three information extraction datasets with standoff annotations (Drug-ADE, CMED, and SHAC), generated predictions were mapped to character offsets by locating each predicted mention in the source document. Predicted and gold annotations were then paired according to the official span and relation matching criteria using a nearest-distance based bipartite matching method, making the results independent of prediction order; in contrast, the official scripts perform a single order-dependent matching pass. Predictions that could not be mapped to the document were counted as false positives. To validate the conversion pipeline, we passed the gold annotations through the same mapping and evaluation procedure, obtaining round-trip scores of 0.9750 entity F1 and 0.9608 relation F1 for Drug-ADE, 0.9479 combined F1 for CMED, and 0.9862 and 0.9872 combined F1 for SHAC-MIMIC and SHAC-UW. Scores below 1.0 reflect information loss during text-to-offset conversion, primarily due to ambiguous mention text, and can limit direct comparison with systems that predict character offsets.

## III. Results

### A. In-Distribution Performance of the Unified Models

Table III showed that the recurrent-depth models were generally competitive with the Qwen3 counterparts on the nine clinical tasks. Compared with Qwen3-1.7B, Ouro-1.4B

Thinking achieved higher performance on 8 of 9 in-distribution datasets, including the largest gain on MedQA (+0.068, 0.5075 versus 0.4399). HRM-Text-1B outperformed Qwen3-1.7B on cohort selection, STS, UMN abbreviations, and CMED (0.005–0.046 improvements), but underperformed on Drug-ADE, Family history, SHAC-MIMIC, MedNLI and MedQA. Compared to Qwen3-8B model, Ouro-2.6B Thinking achieved higher performance on MedQA (+0.039) and STS (+0.012) and underperformed on CMED (-0.017). On the rest 6 of the 9 datasets, the performance differences between the two models are marginal within 0.01.

TABLE III. IN-DISTRIBUTION TEST RESULTS

| Dataset / metric | Qwen3 1.7B | Ouro 1.4B-T | HRM 1B | Qwen3 8B | Ouro 2.6B-T |
|---|---|---|---|---|---|
| Cohort sel., micro-F1 | 0.8458 | 0.9038 | 0.8921 | **0.9125** | 0.9063 |
| Drug-ADE, entity-only F1 | 0.8689 | 0.8715 | 0.8677 | 0.8696 | **0.8771** |
| Drug-ADE, end-to-end F1 | 0.8034 | 0.8063 | 0.7991 | 0.8075 | **0.8170** |
| STS, accuracy | 0.6893 | 0.6966 | **0.7306** | 0.6626 | 0.6748 |
| Family history, F1 | 0.5618 | 0.6126 | 0.5604 | 0.6442 | **0.6534** |
| CMED, trigger F1 | 0.8196 | 0.8437 | 0.8369 | **0.8480** | 0.8321 |
| CMED, attribute F1 | 0.6269 | **0.6818** | 0.6793 | 0.6817 | 0.6600 |
| CMED, combined F1 | 0.7275 | 0.7644 | 0.7609 | **0.7669** | 0.7501 |
| SHAC-MIMIC, trigger F1 | 0.9591 | 0.9610 | 0.9631 | **0.9638** | 0.9631 |
| SHAC-MIMIC, argument F1 | 0.7860 | **0.7939** | 0.7835 | 0.7929 | 0.7888 |
| SHAC-MIMIC, combined F1 | 0.8474 | 0.8531 | 0.8472 | **0.8535** | 0.8505 |
| MedNLI, accuracy | 0.8875 | 0.8987 | 0.8558 | 0.9051 | **0.9128** |
| MedQA, accuracy | 0.4399 | 0.5075 | 0.3857 | 0.5734 | **0.6127** |
| UMN, accuracy | 0.9845 | 0.9826 | 0.9850 | **0.9855** | 0.9833 |

Best result per row in bold. HRM: HRM-Text. Drug-ADE: paragraph-level predictions recombined per note and scored against gold standard annotations (Section II.A).

Across models, the largest performance differences were observed on the knowledge-intensive MedQA benchmark and the family history extraction task, whereas the remaining information extraction tasks, including Drug-ADE, CMED, and SHAC-MIMIC, showed relatively small differences between model families. Overall, the Ouro recurrent-depth models achieved performance comparable to substantially larger decoder-only baselines across most in-distribution tasks, while HRM-Text-1B yielded mixed results.

TABLE IV. OUT-OF-DISTRIBUTION TEST RESULTS

| Dataset / metric | Qwen3 1.7B | Ouro 1.4B-T | HRM 1B | Qwen3 8B | Ouro 2.6B-T |
|---|---|---|---|---|---|
| SHAC-UW, trigger F1 | 0.9481 | 0.9473 | 0.9482 | **0.9514** | 0.9510 |
| SHAC-UW, argument F1 | 0.7628 | 0.7718 | 0.7687 | 0.7750 | **0.7752** |
| SHAC-UW, combined F1 | 0.8294 | 0.8349 | 0.8331 | **0.8383** | **0.8383** |
| NLI4CT, accuracy | 0.4840 | 0.5940 | **0.6300** | 0.5960 | 0.6200 |
| MMLU-Pro, accuracy | 0.3524 | 0.4508 | 0.3123 | 0.5067 | **0.5456** |
| 2010 i2b2, concept F1 | 0.2046 | 0.0980 | 0.1135 | 0.1249 | **0.2233** |

Best result per row in bold; abbreviations as in Table III. SHAC-UW: new institution; NLI4CT: new task format; MMLU-Pro: medical knowledge; i2b2: unseen annotation schema.

### B. *Transfer to Out-of-Distribution Test Sets*

Table IV reported transfer to datasets excluded from fine-tuning. SHAC-UW combined F1 values were closely clustered across models (0.8294–0.8383), with Qwen3-8B and Ouro-2.6B-T tied at 0.8383. On NLI4CT, HRM-Text-1B scored highest (0.6300), followed by Ouro-2.6B-T (0.6200), Qwen3-8B (0.5960), Ouro-1.4B-T (0.5940), and Qwen3-1.7B (0.4840). On MMLU-Pro (Health), Ouro-2.6B-T achieved the highest accuracy (0.5456), whereas HRM-Text-1B scored lowest (0.3123). All models achieved low concept F1 on the 2010 i2b2 schema (0.0980–0.2233)..

### C. *Curriculum Training*

Table V demonstrated mixed effects of the two-stage curriculum training strategy across the five models. CMED combined F1 improved in every model by 0.0075–0.0746, whereas family-history F1 declined by 0.0167–0.0585. SHAC scores generally decreased, with declines approaching 0.03 for Qwen3-1.7B. Four models improved on 2010 i2b2 concept extraction, while HRM-Text-1B declined. These results do not show a consistent benefit across extraction tasks; however, Stage 1 introduced additional training exposure relative to direct fine-tuning, preventing isolation of the curriculum ordering effect.

TABLE V. CHANGE FROM THE TWO-STAGE CURRICULUM

| Dataset | Qwen3 1.7B | Ouro 1.4B-T | HRM 1B | Qwen3 8B | Ouro 2.6B-T |
|---|---|---|---|---|---|
| Drug-ADE entity F1 | −0.0153 | −0.0101 | +0.0021 | −0.0025 | −0.0089 |
| Drug-ADE end-to-end F1 | -0.0168 | -0.0129 | +0.0023 | -0.0036 | -0.0096 |
| Family History F1 | -0.0375 | -0.0585 | -0.0206 | -0.0413 | -0.0167 |
| CMED combined F1 | +0.0746 | +0.0105 | +0.0384 | +0.0179 | +0.0075 |
| SHAC-MIMIC combined F1 | -0.0286 | -0.0202 | +0.0010 | -0.0112 | -0.0084 |
| SHAC-UW combined F1 | -0.0274 | -0.0218 | -0.0052 | -0.0108 | -0.0117 |
| 2010 i2b2 (concept F1) | +0.0084 | +0.0574 | −0.0162 | +0.0152 | +0.0204 |

Abbreviations as in Table III. Two-stage curriculum minus single-stage training under the same fine-tuning configuration.

## IV. DISCUSSION

### A. *Recurrent-Depth Models Compared with Standard Transformers*

In this study, we systematically evaluated two recurrent-depth model families and two standard transformer baselines after joint fine-tuning on 9 clinical tasks. Ouro-1.4B Thinking outperformed Qwen3-1.7B on 11 out of 13 dataset-level comparisons (for extraction datasets we use combined or end-to-end F1 as summary metrics). Ouro-2.6B Thinking performed comparably with Qwen3-8B on most datasets except CMED. These findings are consistent with general-domain evaluations showing that Ouro models can match standard transformers with 2-4 times more parameters [18]. HRM-Text-1B exhibited a mixed profile, scoring higher than Qwen3-1.7B on cohort selection, STS, CMED, UMN abbreviations, and NLI4CT, while scoring lower on MedNLI, MedQA, and MMLU-Pro. Cross-institution generalization remains a challenge in clinical NLP [33]. All five models achieved similar combined F1 on SHAC-UW (0.8294–0.8383), suggesting that transfer between these SHAC institutions was feasible under a shared annotation schema. Performance was substantially lower on 2010 i2b2 concept extraction. Because these benchmarks differ in task formulation, data, and annotation schema, this contrast does not isolate schema shift as the cause of lower performance.

### B. *Comparison with Task-Specific Systems*

We compared our results with published task-specific systems generally developed for individual benchmarks. Across datasets, the best-performing recurrent-depth model achieved competitive performances. Ouro-2.6B Thinking achieved 0.9063 micro-F1 on cohort selection, near the challenge score of 0.9100 [22]. On MedNLI, it achieved 0.9128 accuracy, compared with 0.9020 reported in [34]. For abbreviation disambiguation, our models achieved accuracy values around 0.985 comparable to the published precision of 0.9849 [1]. On SHAC-UW, Ouro-2.6B Thinking achieved 0.8383 combined F1 compared with the challenge score of 0.7740 [5]. On family history, Ouro-2.6B Thinking achieved 0.6534 F1, below the reported 0.6810 [21]. On Drug-ADE, Ouro-2.6B Thinking achieved 0.8771 entity F1 and 0.8170 end-to-end relation F1, compared with 0.9418 and 0.8905 reported in [3]. On SHAC-MIMIC, Ouro-1.4B Thinking achieved the highest combined F1 among recurrent models (0.8531), below the challenge score of 0.9010 [5]. On MedQA, Ouro-2.6B Thinking achieved 0.6127 accuracy; Med-PaLM 2 reported 0.8650 under a different modeling and prompting setup [35]. On the out-of-distribution datasets, task-specific systems achieved higher performance on NLI4CT (0.8560 F1) [27] and 2010 i2b2 concept extraction (0.8996 F1) [34], highlighting remaining challenges in cross-task and cross-schema generalization.

Two factors may contribute to the gap from task-specific extraction systems: (1) our generated mention text must be mapped to character offsets for scoring, whereas many specialized extraction systems use token-level labeling and produce offsets directly; and (2) some previously published systems used additional clinical-domain pretraining [34]. The gold-target round-trip analysis (Section II.E) shows information loss during text-to-offset conversion, although its contribution to the model-level performance gap was not quantified. Richter-Pechanski et al. also reported difficulty generating character offsets with fine-tuned language models [7].

### C. Performance Differences Across Tasks

Across task categories, recurrent-depth models showed different performance patterns relative to standard transformer baselines. Ouro models showed their clearest gains on natural language inference and medical question answering, whereas differences were generally smaller on information extraction tasks. Within information extraction, results varied across datasets and between entity, relation, and attribute prediction. Classification results also varied: HRM-Text-1B improved on cohort selection and STS relative to Qwen3-1.7B, while abbreviation disambiguation accuracy was similar across models. These patterns describe task-level performance differences but do not establish which tasks require more reasoning or why a particular architecture performs better.

These task categories were defined post hoc based on observed performance patterns and should not be interpreted as direct measures of reasoning complexity. For example, abbreviation disambiguation requires contextual interpretation, while extraction tasks include span identification, relation prediction and attribute classification. The stronger performance of Ouro models on inference-oriented tasks is consistent with previous observations that recurrent-depth models improve iterative knowledge manipulation rather than simply increasing knowledge capacity [18]. The weaker performance of HRM-Text-1B on MedQA and MMLU-pro may reflect differences in its training corpus rather than the design of recurrence alone [19].

### D. Implications of Recurrent Computation for Clinical NLP

The observed performance patterns suggest that recurrent-depth models may provide greater benefit for tasks requiring clinical information integration than for tasks focused on explicit extraction. Ouro models showed their largest gains over standard transformers on inference and question-answering tasks, while improvements were limited on span-oriented extraction tasks. However, whether these gains represent improved clinical reasoning remains unclear. Unlike explicit chain-of-thought approaches, recurrent models perform additional computation through hidden-state refinement, and their intermediate "reasoning" processes are not directly observable. Therefore, the observed improvements may reflect more effective iterative representation refinement, better integration of distributed clinical information, or simply increased computation rather than a qualitatively different reasoning capability.

Future studies should compare recurrent depths within the same model family while controlling training data and computational budgets. Comparisons of joint and task-specific fine-tuning would help characterize task interference, and a control matched for total training updates would clarify the contribution of curriculum ordering. Consistent decoding and evaluation pipelines are also needed to separate model errors from output-conversion errors. Finally, analyses of intermediate states and clinician-designed assessments could test whether improved benchmark performance reflects better use of clinical evidence. Explicit reasoning and trajectory-based training are additional directions for investigation [36], [37]. However, meaningful evaluation of these approaches will require new benchmarks that specifically assess medical reasoning, as current benchmarks provide limited supervision and evaluation of clinical reasoning processes.

### E. Differences Between Recurrent Model Families

Ouro and HRM-Text both use recurrent-depth mechanisms that repeatedly refine hidden representations within a forward pass, but their architectures and training histories differ. Ouro repeatedly applies a shared decoder stack, whereas HRM-Text couples high- and low-level recurrent modules that update at different frequencies [18], [19]. Ouro also received substantially more pretraining and additional reasoning-oriented post-training, whereas HRM-Text was pretrained using 40 billion unique tokens over 60 billion processed tokens. These differences may contribute to their distinct task profiles, including HRM-Text's lower scores on knowledge-intensive tasks such as MedQA and MMLU-Pro. More broadly, the results suggest that recurrent computation alone is insufficient to explain performance differences; architecture design and training strategy likely influence how recurrent computation is utilized. These findings motivate future investigations using controlled experimental designs to understand how recurrent architecture, pretraining, and optimization strategies jointly influence model performance in clinical NLP tasks.

## F. Limitations

This study has several limitations. First, we did not control for pretraining: the model families differ in pretraining data and post-training. Consequently, the observed differences cannot be attributed solely to recurrent computation. Second, inference differed across model families. Qwen3 and Ouro used schema-constrained decoding, whereas HRM-Text-1B outputs were generated with PyTorch and validated as JSON without equivalent decoding constraints. Differences in output formatting may therefore have affected performance estimates, particularly for information extraction tasks. Third, we did not compare unified multitask fine-tuning with task-specific fine-tuning, limiting our ability to quantify task interference, nor did we compare direct-answer generation with explicit chain-of-thought prompting. In addition, we evaluated predictive performance but not inference latency, memory usage, or computational cost. Finally, our scoring protocols for STS, NLI4CT, and offset-based extraction (Section II) differ from some prior work, limiting direct comparison with published results.

# V. Conclusion

In this study, we provided the first systematic evaluation of recurrent-depth language models across a diverse set of clinical NLP tasks using a unified multitask fine-tuning framework. Our findings suggest that recurrent-depth architectures can provide an effective approach to improving parameter efficiency while maintaining competitive performance across diverse clinical tasks. At the same time, the contrasting performance of Ouro and HRM-Text suggests that recurrent computation alone is insufficient to explain model performance. Instead, how recurrent architecture is implemented and combined with pretraining and optimization strategies appears to play an equally important role. We hope this work provides a foundation for future studies investigating recurrent-depth architectures for domain-specific language modeling, including how recurrent computation influences knowledge acquisition, reasoning, generalization, and computational efficiency.

# Code and Data Availability

The 2018 n2c2, 2019 n2c2, 2022 n2c2, and 2010 i2b2/VA datasets are available from the n2c2 website after data use agreement (see https://n2c2.dbmi.hms.harvard.edu/ for details). The MedNLI dataset is available from https://physionet.org/content/mednli/1.0.0/. The MedQA-USMLE dataset is available from https://github.com/jind11/MedQA. The UMN abbreviation dataset is available from https://conservancy.umn.edu/handle/11299/137703. The NLI4CT dataset is available from https://sites.google.com/view/nli4ct/home. The MMLU-Pro dataset is available from https://huggingface.co/datasets/TIGER-Lab/MMLU-Pro. The code for data processing, training, and evaluation will be made available at: https://github.com/uf-hobi-informatics-lab/Med-Flan